\documentclass{article}

\usepackage[utf8]{inputenc} 
\usepackage[T1]{fontenc}    
\usepackage{url}            
\usepackage{booktabs}       
\usepackage{amsmath, amssymb, amsfonts}       
\usepackage{microtype}      
\usepackage{graphicx}
\usepackage{tabu}
\usepackage{color}
\usepackage{subcaption}

\title{Filter sharing: Efficient learning of parameters for volumetric convolutions}

\author{
Rahul Venkataramani\thanks{The first three authors contributed equally. Corresponding author: sheshadri.thiruvenkadam@ge.com} , Sheshadri Thiruvenkadam,   Prasad Sudhakar \\ {Hariharan Ravishankar, 
  Vivek Vaidya} \\ \\ GE Global Research, Bangalore, India
}

\def \bs {\boldsymbol}

\begin{document}

\maketitle

\begin{abstract}
Typical convolutional neural networks (CNNs) have several millions of parameters and require a large amount of annotated data to train them. In medical applications where training data is hard to come by, these sophisticated machine learning models are difficult to train. In this paper, we propose a method to reduce the inherent complexity of CNNs during training by exploiting the significant redundancy that is noticed in the learnt CNN filters. Our method relies on finding a small set of filters and mixing coefficients to derive every filter in each convolutional layer at the time of training itself, thereby reducing the number of parameters to be trained. We consider the problem of 3D lung nodule segmentation in CT images and demonstrate the effectiveness of our method in achieving good results with only few training examples. 
\end{abstract}

\section{Introduction}
Deep learning using CNNs are quite successful in learning tasks due to its ability in exploiting spatial context and weight sharing between pixels. With tens of convolutional layers in a typical network, the number of parameters to be learned runs into hundreds of thousands and therefore requires huge amount of training data. In cases of problems involving medical images, annotated data is hard to come by and the dimension of clinical data nowadays, being high-resolution 3D/4D, is huge. Deep learning on such data would require 3D/4D convolutional filters to properly capture the spatial context. In view of above, we run into problems of overfitting and poor generalization, reducing confidence for clinical deployment. Secondly, utilizing large CNN models in devices is also limited by the memory requirements. Often, good CNN models take up several gigabytes of memory and it is hard to deploy on small devices. In this context, it is of great interest to reduce the number of parameters of a CNN model to make it trainable using small data and also amenable for easier deployment. 

It is observed that CNN filters show significant redundancy in the construction of the output maps~\cite{ioannou_2016_53187}. E.g., output maps often use scaled versions of some filters. In this work, we exploit this redundancy to reduce the effective number of parameters without altering the network architecture and performance. We learn only a small set of filters for each layer and then learn how to combine these filters to arrive at the full set of filters. 

 
 

The issue of reducing model size and complexity has been receiving a lot of attention of late. Most of the methods advocate learning full set of filters first and then pruning them. In~\cite{DBLP:journals/corr/KimPYCYS15}, a method for compressing the model parameters is investigated, along with a scheme to fine tune the network to recover some loss incurred due to compression. A similar approach is presented in~\cite{journals/corr/ChenWTWC15a} where the low-frequency nature of the filters is exploited. An elaborate scheme to quantize and Huffman code the parameters is presented in~\cite{DeepCompression}. To efficiently perform the arithmetic operations during the test time,~\cite{courbariaux2014training} uses low-precision fixed point representation of the model parameters. 

A few methods also exploit the low-rank nature of the convolution filters to efficiently perform convolutions and also to reduce the storage requirements~\cite{denton:2014, Jaderberg14b}. A few methods address the problem of directly reducing the number of training parameters.~\cite{ioannou_2016_53187} exploits the separability property of filters to reduce the number of trainable parameters and~\cite{NIPS2013_5025} presents a method to predict the parameters during training by exploiting their the redundancy.  

A work that is in the same spirit as ours is reported in~\cite{Sironi:2014by}. They propose to use the scheme in~\cite{NIPS2013_5025} during training and then perform a decomposition of these learned filters using a small set of filters, whereas we propose to learn the small set of filters and the mixing coefficients during training time itself by performing back-propagation directly on the filter bank and the mixing coefficients.

\section{Learning a Linear decomposition of CNN filters}
Consider a convolutional layer in a deep CNN which takes in $N$ $D$-dimensional feature maps $\bs F_j, 1\leq j\leq N$ and outputs $M$ $D$-dimensional feature maps 
\begin{equation}
\textstyle\bs G_i = \left (\sum_{j=1}^N \bs F_j * \bs v_{ij}  \right ) + \bs b_i,\; 1\leq i \leq M,
\end{equation}
\noindent
where $\bs v_{ij}$ is a filter of size ${S}^{\frac{1}{D}}$ in each of the $D$-dimensions, $*$ is $D$-dimensional convolution and $\bs b_i$ is bias. The CNN learns the filters $\bs v_{ij}$ and the bias $\bs b_i$ by using training data, a loss function and back-propagation algorithm. The number of unknowns to be learned in each layer, sans the bias term, is $MNS$, which can get very large when we have $D>2$. Filter sharing overcomes the burden of large number of parameters by deriving the $MN$ filters using a smaller set of $P\;(\ll MN)$ filters 
\begin{equation}
\textstyle\bs v_{ij} = \sum_{p=1}^P \bs{\widetilde v_{p}} \alpha_{ij}^p ,\; 1\leq i \leq M, 1\leq j \leq N,
\label{filtersharing}
 \end{equation}
where $[\alpha_{ij}^1, \alpha_{ij}^2, \cdots, \alpha_{ij}^P]^T$ is a vector of weights which combines the $P$ filters to produce $\bs v_{ij}$.  

Parameter reduction can also be achieved by reducing the number of filters in each layer. However, this will fail to capture the necessary filter diversity leading to poor performance. Unlike methods which first learn $MN$ filters during training and then trim them down to a smaller set of filters for the test time, we propose to directly learn the filters $\bs{\tilde v}_p$ and the weights $\alpha_{ij}^p$ explicitly during the training time itself. We call this approach as \emph{filter sharing} method and with this approach, the number of parameters to be trained becomes $MNP + PS$. With filter sharing, the number of parameters scales by $P$ when the filter size increases, whereas otherwise it would scale by $M\times N$, which can be very large. We experimentally demonstrate the filter sharing application in the next section.


\section{Experimental results}

In this section, we demonstrate the efficacy of our approach on a challenging 3D medical imaging segmentation problem. We first use CIFAR-10 as a test-bed dataset to demonstrate the usefulness of our method on limited data problems. 

Re-writing the filters $\bs v_{ij}$ as a linear combination of the seed filters $\bs{\widetilde v}_p$, as in Eq.~\eqref{filtersharing} can be seen as an ill-posed matrix factorization problem. Thus, as a means of regularization of the seed filters  $\bs{\tilde v}_p$ and mixing coefficients  $\alpha_{ij}^p$, a variety of schemes were experimented including unit norm constraint on the seed filters, sparsity, rank minimization norms on of $\alpha_{ij}^p$ and dropout on the elements of the output feature maps $\bs G_i$. Dropout regularization with lesser dropout probablity than for regular CNN implementations gave the best results compared to the above possible regularization schemes.
  
\subsection{CIFAR-10}
\begin{figure*}[t]
    \centering
		
		\begin{tabular}{cc}
    \begin{subfigure}[t]{0.49\textwidth}
        \centering
        \includegraphics[height=1.5in]{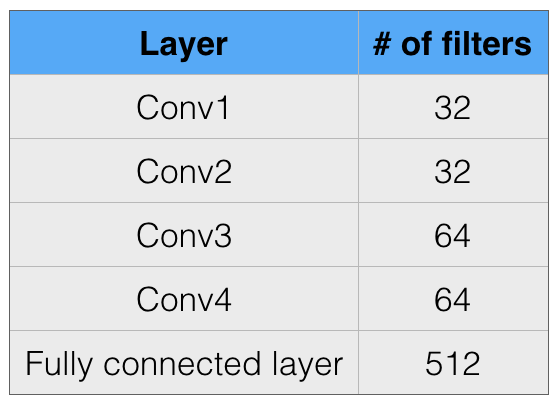}
        \caption{} 
        \label{cifarArchitecture}
    \end{subfigure}
    
    \begin{subfigure}[t]{0.49\textwidth}
        \centering
        \includegraphics[height=1.5in]{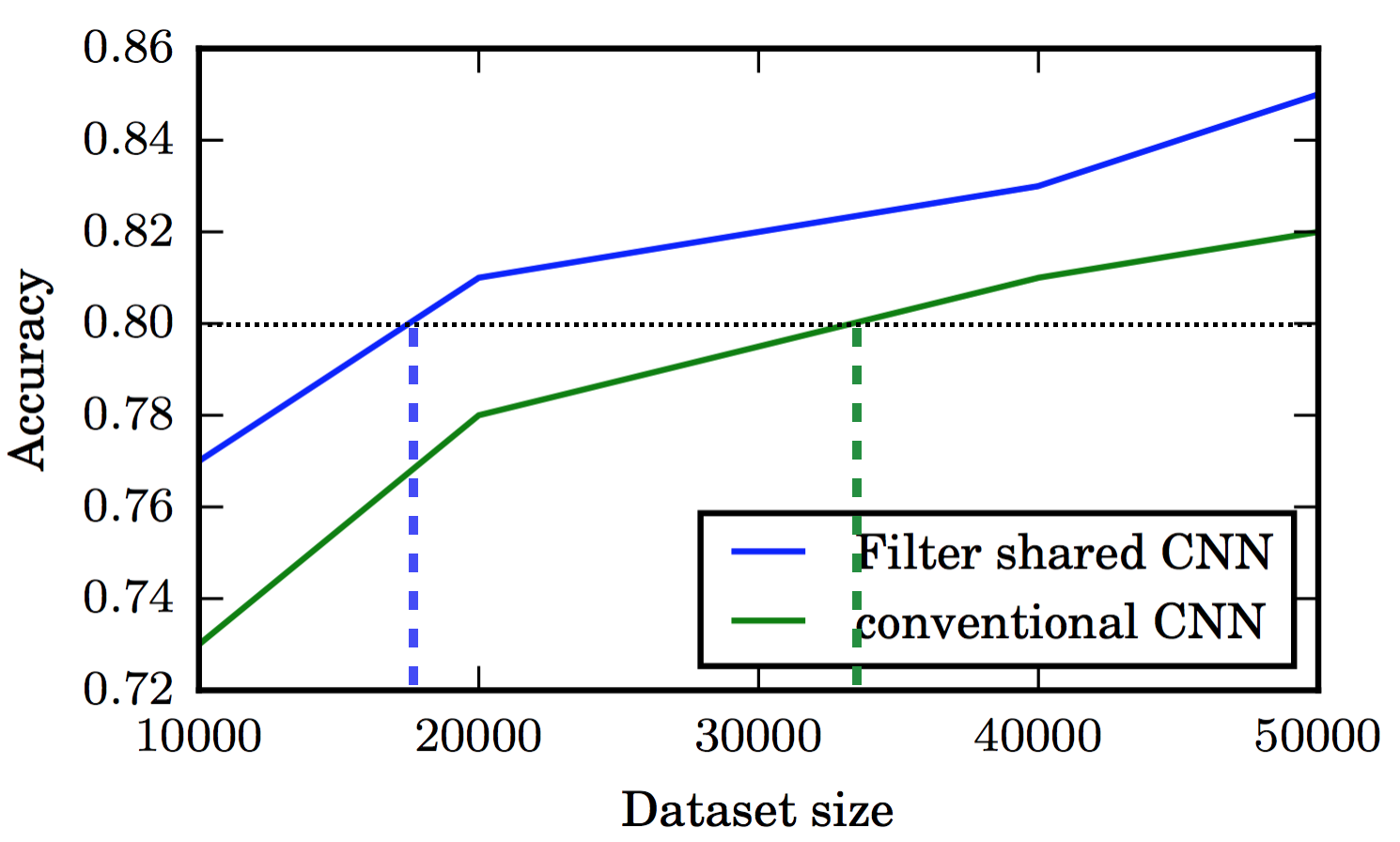}	
        \caption{}
        \label{fig:cifarResults}
    \end{subfigure}
		\end{tabular}
		
		\caption{(\subref{cifarArchitecture}) Exemplar architecture (CIF-CNN) and (\subref{fig:cifarResults}) CIFAR-10 results on validation set. CIF-CNN was used with $P=15$ in each layer for the filter sharing results in (\subref{fig:cifarResults}).}
		\label{fig:cifar}
  \end{figure*}

Using the CIFAR-10 dataset, we wanted to investigate the robustness of proposed approach to limited training data and also to show its applicability on a non-medical, 2-D labeling task. Towards this goal, we define a customized/exemplar architecture shown in Fig.~\ref{cifarArchitecture}. We would like to highlight that the goal of this experiment is not to advance state-of-the-art results on CIFAR-10 dataset but to demonstrate that proposed workflow can be incorporated to any CNN architecture without adversely affecting performance. 
Fig.~\ref{fig:cifarResults} shows the results of the exemplar architecture on validation dataset of $10000$ CIFAR-10 samples learnt on varying sized subsets from $50k$ training samples. Filter sharing achieves an average $\sim 3-4\%$ increase in validation accuracy for all the subsets. Additionally, CIF-CNN-FS achieves $80\%$ validation accuracy using only $\sim 35\%$ of the training dataset ($\sim 17k$ samples) while without filter sharing $\sim33k$ samples are required to achieve the similar performance. 

%
%
\subsection{Lung nodule Segmentation}
Segmentation of relevant structures from 3-D volumetric medical data is critical to morphology measurements for anatomies, characterization of lesions and other applications. We have chosen lung nodule segmentation from 3-D Low Dose CT (LDCT) images as an application for the proposed filter sharing method due to its clinical relevance and challenging nature. We work with 3-D LDCT volumes from Lung Image Database Consortium~\cite{armato2011lung}. The LIDC-IDRI database consists of data from 1,010 patients along with annotation by four experienced radiologists. We pre-selected 93 large solid data with $\sim 280$ lesions, of which $50\%$ of lesions ($\sim 140$) were used for training, and remaining $50\%$ was equally split for validation and testing purpose.

\begin{figure*}[b]
    \centering
		
    \begin{subfigure}{0.51\textwidth}
        \centering
        \includegraphics[scale=0.3]{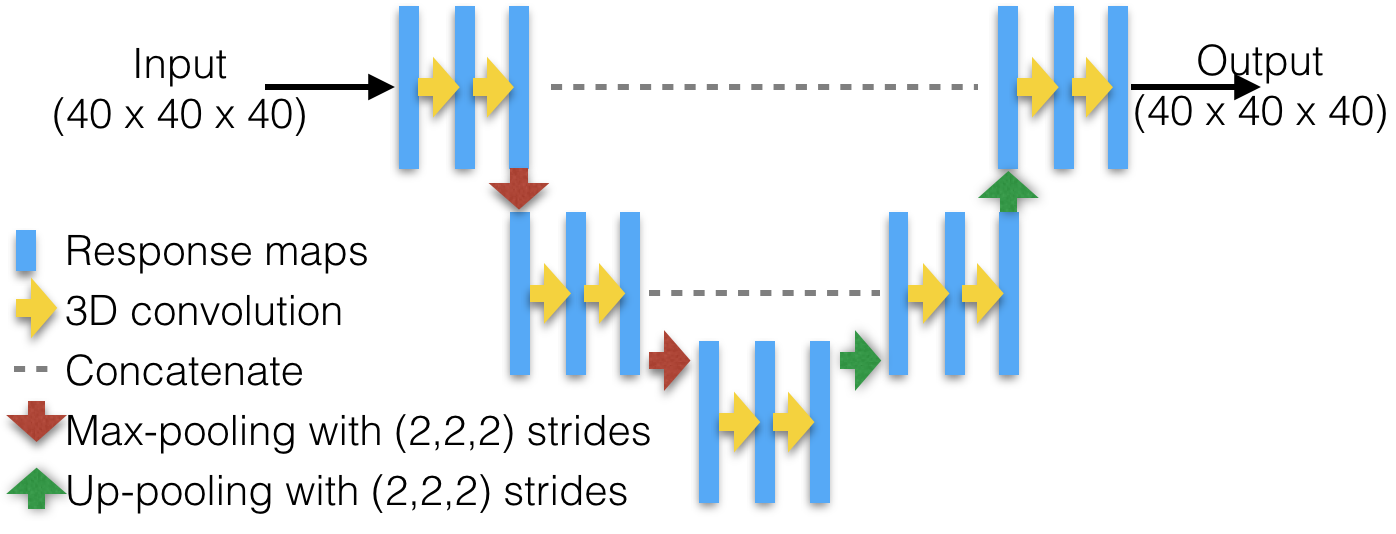}
	\caption{} 
        \label{fig:UNet}
    \end{subfigure}  
    \begin{subfigure}{0.45\textwidth}
        \centering
        \includegraphics[scale=0.65]{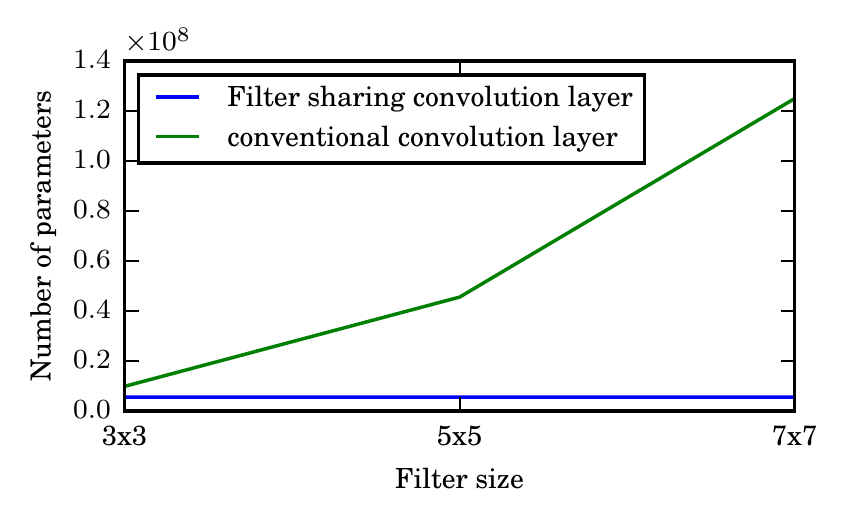}	
        \caption{}
        \label{fig:params}
    \end{subfigure}
		
		\caption{(\subref{fig:UNet}) 3D U-Net architecture and (\subref{fig:params}) Number of parameters to be trained as a function of the size of convolution filter for 3D U-Net architecture}
		\label{fig:cifar}
  \end{figure*}

Fully convolutional networks (FCN) have recently emerged as state of the art algorithms for natural and medical image segmentation, multi-object scene labeling and related tasks. Additionally, FCNs have analysis-synthesis arms to produce output volumes same/similar to the size of input volumes leading to roughly double the number of parameters compared to a standard CNN label prediction network, making them an ideal candidate to demonstrate the proposed method's usage.  Fig.~\ref{fig:UNet} shows 3-D U-Net architecture we have used in our experiments~\cite{cciccek20163d}. Fig.~\ref{fig:params} shows how the number of parameters in a 3D U-Net convolution layer scales with respect to the size of convolutional filter. As the size of the filters increases, filter sharing method clearly shows its advantage. We would like to highlight that our task is segmentation of nodules and not detection which we assume is done automatically or manually. Input to the architecture are $40\times 40\times 40$ LDCT volumes pre-identified as containing nodules and output are the segmentation masks predicted by the network.

We use Dice overlap between the predicted nodule segmentation and the ground truth annotation (provided by an expert) as the metric to evaluate performance. Table ~\ref{U-NetTable} contains the comparisons. We achieve a significant $6\%$ increase in average Dice overlap, in addition to achieving $45\%$ reduction in parameters using FS. We hypothesize that learning lesser parameters in limited data scenarios, like medical imaging problems, can lead to more efficient, stable and better architectures. 

{\setlength{\extrarowheight}{5pt}
\begin{table}[]

\centering
  \begin{tabu}{| c | c | c |}
    \hline
		
    \rowfont[c]{\bfseries}   Architecture & Number of parameters & Dice overlap in \%\\ \hline

    3D U-Net without FS & 9847689 & 60 \\ \hline
		
	3D U-Net with FS & \bfseries 5505572 & \bfseries 66 \\
    \hline
  \end{tabu}
	\vspace{0.2cm}
  \caption{Effect of using filter shared 3D convolutional layers in U-Net}
  \label{U-NetTable}
\end{table}}

\section{Discussion}
In this paper, we have proposed a method to reduce the number of parameters to be learned in a CNN by linearly expanding each convolutional filter in terms of a small set of filters. This method has great advantages when training data is small to avoid overfitting. Moreover, reduction in the number of parameters also helps in reducing the overall memory footprint of the model, making it amenable for deployment in diverse situations. 

With filter sharing framework, we can also impose regularization on the kind of filters to be learnt. Also, we can include a fixed set of standard filters (e.g., wavelet filters) to improve the performance. Learning a latent set of filters also helps in easily transferring models across different deep learning architectures. This is extremely relevant when we would like to arbitrarily increase or decrease the number of filters in a given layer of a deep network. These issues are a part of our future work. 

\bibliographystyle{plain}
\bibliography{biblio}

\end{document}